\documentclass[lettersize,journal]{IEEEtran}
\usepackage{amsmath,amsfonts}
\usepackage{algorithmic}
\usepackage{algorithm}
\usepackage{array}
\usepackage[caption=false,font=normalsize,labelfont=sf,textfont=sf]{subfig}
\usepackage{textcomp}
\usepackage{stfloats}
\usepackage{url}
\usepackage{verbatim}
\usepackage{graphicx}
\usepackage{cite}
\usepackage{svg}
\usepackage{bbding} 
\usepackage{mathrsfs}
\usepackage{booktabs}

\usepackage{makecell}
\usepackage{float}
\usepackage{color}
\usepackage{dsfont}
\usepackage{bbm}
\usepackage{multirow}

\usepackage{threeparttable}

\begin{document}


\title{Robust Image Stitching with Optimal Plane}

\author{Lang Nie,~\IEEEmembership{Member,~IEEE}, Yuan Mei, Kang Liao,~\IEEEmembership{Member,~IEEE}, Yunqiu Xu,~\IEEEmembership{Member,~IEEE}, \\Chunyu Lin,~\IEEEmembership{Member,~IEEE}, Bin Xiao,~\IEEEmembership{Senior Member,~IEEE}
		
\thanks{Lang Nie and Bin Xiao are with the School of Artificial Intelligence, Chongqing University of Posts and Telecommunications, Chongqing 400065, China (e-mail: nielang@cqupt.edu.cn, xiaobin@cqupt.edu.cn). }
\thanks{Yuan Mei is with the Department of Aeronautical and Aviation Engineering, The Hong Kong Polytechnic University, Hong Kong, China (e-mail: yyuan.mei@connect.polyu.hk). }
\thanks{Kang Liao is with the School of Computing and Data Science, Nanyang Technological University, Singapore (e-mail:  kang.liao@ntu.edu.sg).}
\thanks{Yunqiu Xu is with the College of Computer Science and Technology, Zhejiang University, Zhejiang 310058, China (e-mail: imyunqiuxu@gmail.com). }
\thanks{Chunyu Lin is with the Institute of Information Science, Beijing Jiaotong University, Beijing 100044, China (e-mail: cylin@bjtu.edu.cn).}
 \thanks{This work was supported by the National Natural Science Foundation of China (NSFC) under Grants 62502057, 62536002, and 62402432. (The first two authors contribute equally. Corresponding author: Bin Xiao.)\\
 }
 }

\markboth{Journal of \LaTeX\ Class Files,~Vol.~14, No.~8, August~2021}%
{Shell \MakeLowercase{\textit{et al.}}: A Sample Article Using IEEEtran.cls for IEEE Journals}


\maketitle

\begin{abstract}
\label{sec:Abstrat}
We present \textit{RopStitch}, an unsupervised deep image stitching framework with both robustness and naturalness.
To ensure the robustness of \textit{RopStitch}, we propose to incorporate the universal prior of content perception into the image stitching model by a dual-branch architecture. It separately captures coarse and fine features and integrates them to achieve highly generalizable performance across diverse unseen real-world scenes. Concretely, the dual-branch model consists of a pretrained branch to capture semantically invariant representations and a learnable branch to extract fine-grained discriminative features, which are then merged into a whole by a controllable factor at the correlation level.
Besides, considering that content alignment and structural preservation are often contradictory to each other, we propose a concept of virtual optimal planes to relieve this conflict. To this end, we model this problem as a process of estimating homography decomposition coefficients, and design an iterative coefficient predictor and minimal semantic distortion constraint to identify the optimal plane. 
This scheme is finally incorporated into \textit{RopStitch} by warping both views onto the optimal plane bidirectionally. 
Extensive experiments across various datasets demonstrate that \textit{RopStitch} significantly outperforms existing methods, particularly in scene robustness and content naturalness. The code is available at {\color{red}\url{https://github.com/MmelodYy/RopStitch}}.
\end{abstract}
		
\begin{IEEEkeywords}
			Image stitching, Homography, Thin-plate spine 
\end{IEEEkeywords}
 


\section{Introduction}

Image stitching, a long-standing task~\cite{brown2007automatic} in computer vision, aims to generate a wide Field-of-View (FoV) panorama from multiple images with limited FoV, while minimizing content artifacts and shape distortions. Besides enabling user creation and entertainment, it has been widely applied in various fields such as virtual reality~\cite{rhee2017mr360, danyluk2020touch}, autonomous driving~\cite{hou2021visual}, and intelligent surveillance.

\begin{figure}[t]
  \centering
  \includegraphics[width=\linewidth]{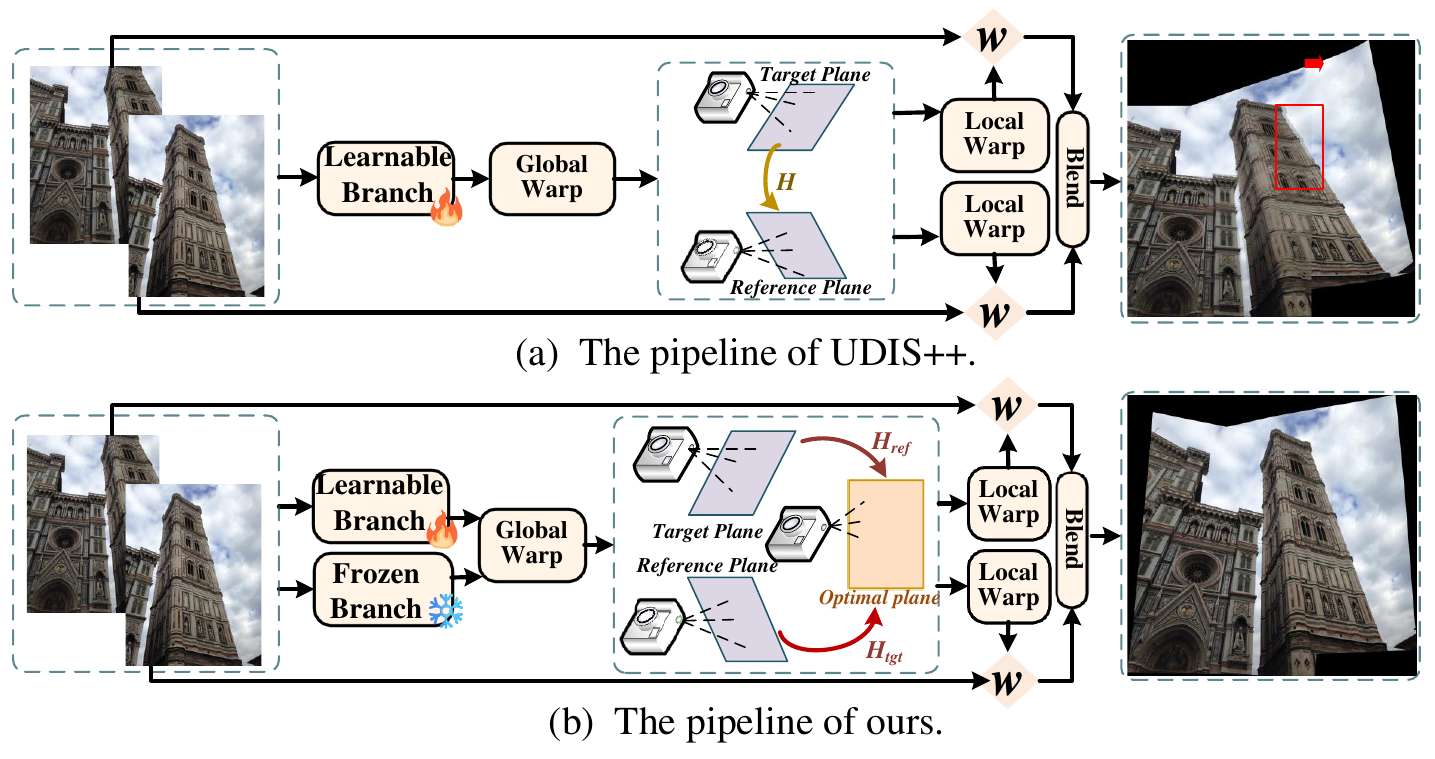} 
  \caption{Difference from existing solutions. (a) UIDS++ \cite{nie2023parallax} uses a single-branch architecture and a single reference plane, resulting in limited cross-scene generalization and content stretching. (b) Our method incorporates perceptual prior through a dual-branch architecture and aligns images on a virtual optimal plane, thereby enhancing cross-scene generalization and natural appearance.} 
  \label{fig:pipeline}
\end{figure}

Traditional image stitching algorithms have achieved considerable success in both content alignment~\cite{zaragoza2013projective} and shape preservation~\cite{chen2016natural}, but they typically rely on an implicit assumption, \textit{i.e.}, the presence of texture-rich scenes with distinct features. When this assumption is not satisfied, these traditional algorithms can cause performance degradation or even failure, especially in low-texture or low-light scenes, where hand-crafted features (\textit{e.g.}, keypoint~\cite{lowe2004distinctive}, line segment~\cite{von2008lsd}, \textit{etc.}) are not sufficiently and evenly detected.

In contrast, deep image stitching solutions get rid of this limitation by digging into high-level semantic features~\cite{nie2021unsupervised} instead of shallow geometric cues, whose capability is acquired through fitting abundant data to enable general semantic perception. However, limited by the scale of existing image stitching datasets (\textit{e.g.}, UDIS-D~\cite{nie2021unsupervised} only contains about 10k samples), the pre-trained models struggle to achieve highly generalizable performance across diverse unseen real-world scenarios due to the domain distribution gap between different datasets.

To overcome the above issues, as shown in Fig. \ref{fig:pipeline}, we propose to incorporate the universal prior (acquired by large-scale datasets) into the image stitching model (trained on limited data quantity). To this end, we present a robust image stitching model with a dual-branch architecture. The first branch leverages a pretrained backbone from large-scale datasets~\cite{deng2009imagenet} to capture semantically invariant features, which implicitly contain the prior of coarse but robust content perception. The other branch adopts a learnable backbone to extract fine-grained discriminative features, which serves as a perfect complement to provide fine but dedicated semantics. 
To combine the advantages of different branches, we carry out the feature aggregation process in the correlation layer, which first calculates respective correlation volumes and then merges them by a controllable factor. The merged correlation is used to predict stable warping parameters and realize robust image stitching.

Furthermore, we propose to relieve the conflict between content alignment and shape preservation by identifying an optimal stitching plane. It decreases the distortion by redistributing the warping burdens from a single view to different views according to a principle of minimal semantic distortions. Specifically, we reformulate the process of determining the optimal plane as generating a set of coefficients for homography decomposition. To find the desired coefficients, we design an iterative coefficient predictor to disentangle a given homography matrix into two and a corresponding semantic distortion constraint to comprehensively measure the deformation distortion and semantic significance. The stitching naturalness can be increased through projecting different views onto the optimal plane bidirectionally.

With the proposed dual-branch architecture and optimal plane, we construct a new unsupervised image stitching framework, named \textit{RopStitch}, which demonstrates superior performance, especially in robustness and naturalness. Our main contributions are centered around:

\begin{itemize}
    \item To improve robustness in real-world scenes, we propose to integrate the universal prior into the image stitching model through a dual-branch architecture.
    \item To decrease structure distortions, we present to determine an optimal stitching plane at the principle of minimal semantic distortions without sacrifying content alignment.
    \item The proposed framework, \textit{RopStitch}, outperforms existing solutions, particularly in scene robustness and content naturalness.
\end{itemize}

\section{Related work}
\label{gen_inst}
\subsection{Traditional Image Stitching}
Traditional image stitching solutions typically leverage hand-crafted features to estimate the desired warps. Based on their optimization objectives, these methods are broadly categorized into two groups: (1) alignment-oriented approaches, and (2) structure-preserving approaches. The first category prioritizes precise content alignment across overlapping regions. Early techniques often relied on global homography models~\cite{brown2007automatic}, which are effective for planar scenes but fail under significant parallax. To handle more complex scenarios, subsequent works introduced more flexible deformation models, such as mesh-based warps~\cite{zaragoza2013projective}, which allow local adjustments; thin-plate splines (TPS)~\cite{li2017parallax} for smooth deformation fields; triangular meshes~\cite{li2019local,cai2025object} to accommodate non-planar surfaces; and even subpixel-level warping techniques~\cite{lee2020warping} for higher precision. To improve alignment quality, some methods incorporate additional cues such as semantic masks~\cite{li2021image,liao2025parallax} and depth maps~\cite{liao2022natural} to identify and suppress outliers in non-coplanar regions.  But excessive alignment often leads to structural distortion, especially in large parallax areas.

In contrast, structure-preserving approaches emphasize maintaining the geometric integrity of salient objects and linear structures in the scene. They use diverse geometric features beyond keypoints, including line segments~\cite{von2008lsd}, edges~\cite{du2022geometric}, and object contours~\cite{cai2025object}. By incorporating these features into the optimization process, they aim to produce warps that minimize bending and stretching of important visual elements~\cite{liao2019single,zhang2020content,jia2021leveraging,chen2016natural,lin2015adaptive,li2024face}. Nevertheless, enforcing structural constraints often comes at the cost of reduced alignment accuracy, which can introduce visible misalignment or ghosting artifacts in overlapping areas.

Moreover, a common challenge of these solutions is their dependence on the quality and quantity of hand-crafted features. The stitching performance could easily degrade in scenes with weak edges, low texture, or lacking clearly defined structures.

\subsection{Deep Image Stitching}
In recent years, deep learning-based image stitching methods have emerged as a promising alternative, offering great robustness and efficiency. Compared to traditional approaches, these methods benefit from the ability of convolutional neural networks (CNNs) to extract high-level semantic features, which are often more invariant to appearance changes and noise. Deep stitching methods mirror the taxonomy of traditional techniques, including homography-based models~\cite{mei2024dunhuangstitch,jiang2022towards,ma2023overview} that regress global projective parameters; mesh-based methods~\cite{nie2023parallax,nie2025stabstitch++,zhang2024recstitchnet} that predict vertex displacements for flexible local warping; and pixel-level approaches~\cite{jia2023learning,kweon2023pixel} that generate dense deformation fields. 

Besides, another significant trend is the shift in learning paradigms aimed at improving generalization. Early methods relied on fully supervised training using ground-truth warps~\cite{nie2020view,nie2022learning,dai2021edge,edstedt2024roma}, which limited their scalability due to the difficulty of labeling real-world data. To address this, researchers developed weakly-supervised frameworks~\cite{song2022weakly} in some specific panoramic imaging setups. More recently, there has been a growing focus on unsupervised learning techniques~\cite{nie2021unsupervised,nie2023parallax,nie2025stabstitch++,jia2023learning,jiang2022towards,jiang2024multispectral}, which train networks using only the input image pairs without any labeled data. 

Despite these advances, even the most robust unsupervised methods inevitably face the domain gap challenge between the training data and unseen cross-scene data. This persistent generalization issue remains the most pressing challenge for unsupervised approaches.
	\section{Methodology}
\label{headings}

\begin{figure*}[t]
  \centering
  \includegraphics[width=0.98\textwidth]{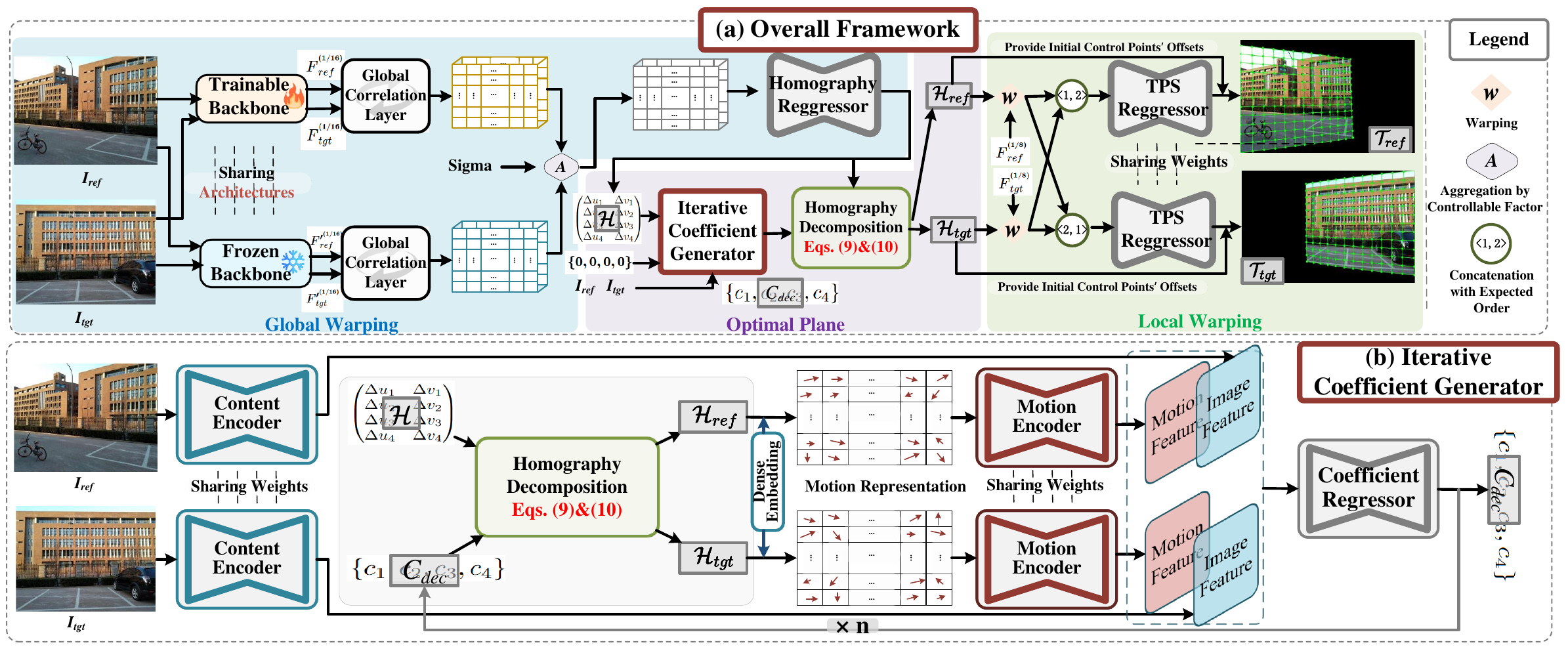} 
  \caption{The framework of \textit{RopStitch}. It takes a dual-branch architecture to construct a robust correlation volume, thereby ensuring robust global transformation. Then the single-view projection is decomposed into two bidirectional warps on the optimal plane, which is followed by the bidirectional local deformation.} 
  \label{fig:framework}
\end{figure*}

\subsection{Overview}
\textit{RopStitch} employs a global-to-local warping strategy, which is similar to UDIS++~\cite{nie2023parallax}. In the global stage, we estimate the homography as the global transformation, while in the local stage, we adopt mesh deformation specified by TPS control points~\cite{bookstein1989principal}. The main difference lies in two aspects: dual-branch architecture and virtual optimal plane. 
As shown in Fig. \ref{fig:framework}(top), given a pair of reference and target images ($I_{ref}$ and $I_{tgt}$) with the resolution of $H\times W$, the dual-branch architecture outputs a global homography $\mathcal{H}$, which contains an implicit universal prior and enables robust estimation. Then $\mathcal{H}$ is decomposed into $\mathcal{H}_{ref}$ and $\mathcal{H}_{tgt}$ to identify the optimal plane. Subsequently, the local stage predicts respective TPS transformations ($\mathcal{T}_{ref}$ and $\mathcal{T}_{tgt}$), carrying out bidirectional elastic warping on the desired plane.

As for the objective function, it consists of three terms as follows:
\begin{equation}
    \mathcal{L} = \mathcal{L}_{align} + w_{s}\mathcal{L}_{shape} + w_{c}\mathcal{L}_{coef},
    \label{eq:sdl}
\end{equation}
where the weights $w_{s}$ and $w_{c}$ are set to 4 and 10. $\mathcal{L}_{align}$ and $\mathcal{L}_{shape}$ constrain the content alignment and shape preservation, and are discussed in Sec. \ref{sec:branch}. $\mathcal{L}_{coef}$ is the constraint to determine an optimal plane, which is described in detail in Sec. \ref{sec:plane}.

\subsection{Dual-branch Architecture}
\label{sec:branch}
Generalizing across diverse scenes is a major bottleneck for stitching algorithms. Fortunately, large-scale pre-trained models often inherently possess the capability of cross-scene understanding. 
To leverage such universal priors, we constructed a dual-branch architecture designed to separately represent learnable features and invariant features from the raw input images. Specifically, we built a Siamese backbone network that was pre-trained on the ImageNet dataset. These two backbones ($f_{frozen}(\cdot)$ and $f_{train}(\cdot)$) share an identical network structure; the sole distinction is that one backbone remains entirely frozen throughout the training process, while the other is kept active (trainable). Furthermore, a potential advantage of the Siamese network structure is its capacity to effectively alleviate discrepancies in feature representations between the two streams, which positively advances the subsequent motion regression and facilitates the cross-domain robustness, as validated in Table \ref{tab3}.

Concretely, the two branches, $f_{frozen}(\cdot)$ and $f_{train}(\cdot)$, extract features at $1/8$ and $1/16$ scale from the input pair $(I_{ref}$ and $I_{tgt})$. We describe this process as the formula below.

\begin{equation}
\begin{aligned}
   F^{(1/8)}_{ref}, F^{(1/16)}_{ref};F^{(1/8)}_{tgt}, F^{(1/16)}_{tgt} = f_{train}(I_{ref};I_{tgt}),\\
    F^{'(1/16)}_{ref}; F^{'(1/16)}_{tgt} = f_{frozen}(I_{ref};I_{tgt}).
   \end{aligned}
\end{equation}





\subsubsection{Correlation-wise Aggregation}
With the introduction of a frozen backbone, we are able to extract invariant feature representations from the input images. However, the optimal usage of these invariant features remains an open research question. The most straightforward approach is to directly add or concatenate the features. However, such methods often lead to convergence between the learnable and invariant features, resulting in redundant feature representations. Similarly, dynamic feature fusion strategies may cause the fusion mechanism to become overly dependent on the training dataset, thereby limiting the model’s generalization ability in cross-domain scenarios.

To address these challenges, we propose to shift the fusion process to the correlation level. After computing the global correlation using CCL~\cite{9605632}, the resulting volumes ($Corr_{train}$,$Corr_{frozen}\in \mathbb{R}^{2\times \frac{H}{16}\times \frac{W}{16}}$) effectively capture the dense spatial correspondences between the reference and target images. The specific formula is as follows:

\begin{equation}
\begin{aligned}
   Corr_{train} = CCL(F^{(1/16)}_{ref},F^{(1/16)}_{tgt}), \\
    Corr_{frozen} = CCL(F^{'(1/16)}_{ref},F^{'(1/16)}_{tgt}).
   \end{aligned}
\end{equation}

Subsequently, by performing aggregation at the level of the global correlation map, we can circumvent the aforementioned issues and achieve more effective integration of invariant features and learnable features. However, how to fuse these features in a manner that simultaneously enhances the robustness of the learnable backbone and preserves the representational stability of the frozen backbone remains a problem to be solved. Inspired by the Variational Autoencoder (VAE) framework \cite{kingma2013auto}, we employ a \textbf{random} $\sigma$ value ($\sigma\in \mathbb{R}^{2\times 1\times 1}$) within the range $[0,1]$  during training to fuse the features, analogous to the roles of variance and mean in VAEs. This process can be written as:
\begin{equation}
   Corr_{fusion} = (1 - \sigma) \cdot Corr_{train} + \sigma \cdot Corr_{frozen}.
\end{equation}
The introduction of random $\sigma$ extends the model’s capacity to represent global correlation maps of varying strengths, rather than being restricted to the optimal correlation patterns present in the training data. 

To fully exploit the prior knowledge encoded in the frozen backbone, we employ a ternary search strategy during inference to adaptively adjust $\sigma$ within the range of $[-1,2]$. Doubling the search outward is equivalent to performing a complementary operation of residual relevance. 

Finally, the $Corr_{fusion}$ is fed into a subsequent global regression network to predict the four-point offsets and further converted into the matrix representation~\cite{cai2025fast}. 
It simultaneously integrates universal content understanding from the frozen branch and fine-grained discrimination from the trainable branch, thus ensuring robust cross-scene stitching capability.

\subsubsection{Fundamental Loss Function} For the sake of method completeness, we briefly describe the basic loss of \textit{RopStitch}, which consists of an alignment term and a shape term.

\textbf{The alignment term} is divided into two parts: global alignment using homography and local alignment using TPS. The specific formulas are as follows:

\begin{equation}
  \begin{matrix}
    \begin{aligned}
    \mathcal{L}_{align} = & \lambda_\mathcal{H}||\mathcal{W}(I_{ref}, \mathcal{H}_{ref}) - \mathcal{W}(I_{tgt}, \mathcal{H}_{tgt} )||_{1}\odot M_{\mathcal{H}} +   \\
    & \lambda_\mathcal{T}||\mathcal{W}(I_{ref}, \mathcal{T}_{ref}) - \mathcal{W}(I_{tgt}, \mathcal{T}_{tgt} )||_{1}\odot M_{\mathcal{T}},   \\
    where& \quad M_\mathcal{H} = \mathcal{W}(\mathbb{I}, \mathcal{H}_{ref})\odot\mathcal{W}(\mathbb{I}, \mathcal{H}_{tgt}),\\ & \quad M_\mathcal{\mathcal{T}} = \mathcal{W}(\mathbb{I}, \mathcal{T}_{ref})\odot\mathcal{W}(\mathbb{I}, T_{tgt}).\\ 
  
    \end{aligned}
  \end{matrix}
\end{equation}
$\mathcal{W}(\cdot,\cdot)$ is the warping operation and $\mathbb{I}$ is an all-one matrix with the same resolution as input images. The weights $\lambda_\mathcal{H}$ and $\lambda_\mathcal{T}$ are set to 1 and 4 to balance global and local warping. The masks $M_\mathcal{H}$ and $M_\mathcal{\mathcal{T}}$ identify overlapping areas, with a value of 1 for overlap and 0 otherwise. $\odot$ is the element-wise multiplication.
\vspace{0.2cm}

\textbf{The shape term} includes an intra-grid constraint and an inter-grid constraint as:
\begin{equation}
	\mathcal{L}_{shape} = \mathcal{L}_{shape}^{intra} + \mathcal{L}_{shape}^{inter}. 
\end{equation}
This allows for a reasonable magnitude of image warping and ensures consistent motions of neighboring control points. Concretely, we define the TPS control points evenly distributed on the whole image with $(U+1)\times(V+1)$ points. These control points can be regarded as the vertices of a $U\times V$ mesh. For brevity, we only describe the constraints for one mesh, even though two meshes are generated from two views.

\vspace{0.1cm}
\noindent \textit{Intra-grid Constraint}:
To prevent the mesh from excessive deformation, we set a minimum length and width in both horizontal and vertical directions, which restricts the deformed mesh from being too small and thereby causing dramatic scaling. Denoting $\vec{i}$ and $\vec{j}$ as the unit vectors in the horizontal and vertical directions, we can formulate this constraint as:
\begin{equation}
  \begin{matrix}
    \begin{aligned}
      \mathcal{L}_{shape}^{intra} = &\frac{1}{^{(U+1) \times V}} \sum_{\{e_{h}\}}{L_{h}^{intra}} + \frac{1}{^{U \times (V+1)}} \sum_{\{e_{v}\}}{L_{v}^{intra}}, \\
      where \quad  &\mathcal{L}_{h}^{intra} = ReLU(\alpha \frac{_W}{^V}-\begin{vmatrix}\vec{e_{h}}\cdot\vec{i}\end{vmatrix}),\\
      &\mathcal{L}_{v}^{intra} = ReLU(\alpha \frac{_H}{^U}-\begin{vmatrix}\vec{e_{v}}\cdot\vec{j}\end{vmatrix}). 
    \end{aligned}
  \end{matrix}
\end{equation}
$\vec{e_{h}}$ and $\vec{e_{v}}$ represent the direction vectors connected by neighboring mesh vertices in the horizontal and vertical directions. The hyperparameter $\alpha$ defines the minimum grid length ratio as a constraint for grid deformation, which is set to 1/8.
\vspace{0.1cm}

\noindent \textit{Inter-grid Constraint}:
To prevent straight structures from bending, the neighboring grids should undergo similar transformations.
Given consecutive edges $\vec{e_{t_{1}}}$ and $\vec{e_{t_{2}}}$ in two neighboring grids, the inter-grid loss can be expressed as follows:

\begin{equation}
	\mathcal{L}_{shape}^{inter} =\frac{1}{K}\sum_{\{\vec{e_{t_{1}}},\vec{e_{t_{2}}}\}}{( 1 - \frac{\begin{vmatrix}\vec{e_{t_{1}}}\cdot\vec{e_{t_{2}}}\end{vmatrix}}{\begin{vmatrix}\vec{e_{t_{1}}}\end{vmatrix} \cdot \begin{vmatrix}\vec{e_{t_{2}}}\end{vmatrix}} )},
\end{equation}
where $\textit{K}$ denotes the number of tuples of two consecutive edges in the mesh.

\subsection{Warping on Optimal Plane}
\label{sec:plane}
Existing image stitching methods typically warp one image to align with the other. However, this strategy places the entire warping burden on a single view, which could easily yield excessive deformation and cause projective distortions. To this end, we propose the concept of optimal stitching planes and carry out bidirectional warping.

Concretely, we use homography $\mathcal{H}$ to represent the plane and decompose this transformation into two ($\mathcal{H}_{ref}$, $\mathcal{H}_{tgt}$) to denote the warps from respective planes to the optimal plane. Denoting $\mathcal{H}$ with 4-pt representation~\cite{detone2016deep}, we can define $\mathcal{H}_{tgt}$ with decomposition coefficients $C_{dec}=\{c_{1}, c_{2}, c_{3}, c_{4}\}$, in which $c_{i}\in[0,1]$ corresponds to one of the four vertexes. This process can be written as:

\begin{equation}
\mathcal{H} \sim \begin{pmatrix}
 \Delta u_1& \Delta v_1 \\
 \Delta u_2& \Delta v_2 \\
 \Delta u_3& \Delta v_3\\
\Delta u_4& \Delta v_4
\end{pmatrix}, 
\mathcal{H}_{tgt}\sim \begin{pmatrix}
 \Delta c_1u_1& \Delta c_1v_1 \\
 \Delta c_2u_2& \Delta c_2v_2 \\
 \Delta c_3u_3& \Delta c_3v_3\\
\Delta c_4u_4& \Delta c_4v_4
\end{pmatrix},
\end{equation}
where $(\Delta u_i,\Delta v_i)$ are the offsets of $i$-th vertex. Then we can derive $\mathcal{H}_{ref}$ as follows:
\begin{equation}
\mathcal{H}_{ref} = \mathcal{H}^{-1}\mathcal{H}_{tgt}.
\end{equation}

Apparently, the key to determining the optimal plane is to precisely identify the decomposition coefficients. To this end, we design an iterative coefficient generator and the corresponding objective function to ensure minimal semantic distortion.

\subsubsection{Iterative Coefficient Generator} As illustrated in Fig. \ref{fig:framework}(bottom), the iterative coefficient generator consists of three sub-networks: content encoder, motion encoder, and coefficient reggressor. The content encoder takes two views as input and encodes them into latent image features, while the motion encoder inputs dense motion representations and encodes them into latent motion features.
Given a $3\times 3$ homography matrix $\mathcal{H} = (h_1\ h_2\ h_3)^T$, where $h_i \in \mathbb{R}^{3 \times 1}$ is a column vector (so $h_i^T$ is the $i$-th row), we embed it into a pixel-wise motion representation as:
\begin{equation}
    m_{x,y}=(\frac{h_1^T\cdot(x\ y\ 1)^T}{h_3^T\cdot(x\ y\ 1)^T}-x,\ \frac{h_2^T\cdot(x\ y\ 1)^T}{h_3^T\cdot(x\ y\ 1)^T}-y),
    \label{eq2}
 \end{equation}
where $(x,y)$ denotes the pixel location in an image.
After acquiring the image and motion features, a coefficient regressor is leveraged to aggregate these features and generate our desired decomposition coefficients $C_{dec}$. Then $C_{dec}$ is input into the homography decomposition module with the predicted homography, which will be decomposed into two bidirectional transformations. 

With $C_{dec}$ updating, the motion representations will be updated accordingly, which formulates an iterative process. For the first iteration, we set $C_{dec}=\{0,0,0,0\}$. This setting makes $\mathcal{H}_{tgt}$ an identity matrix, meaning the initial virtual plane coincides with the reference plane.

\begin{figure}[t]
  \centering
  \includegraphics[width=0.42\textwidth]{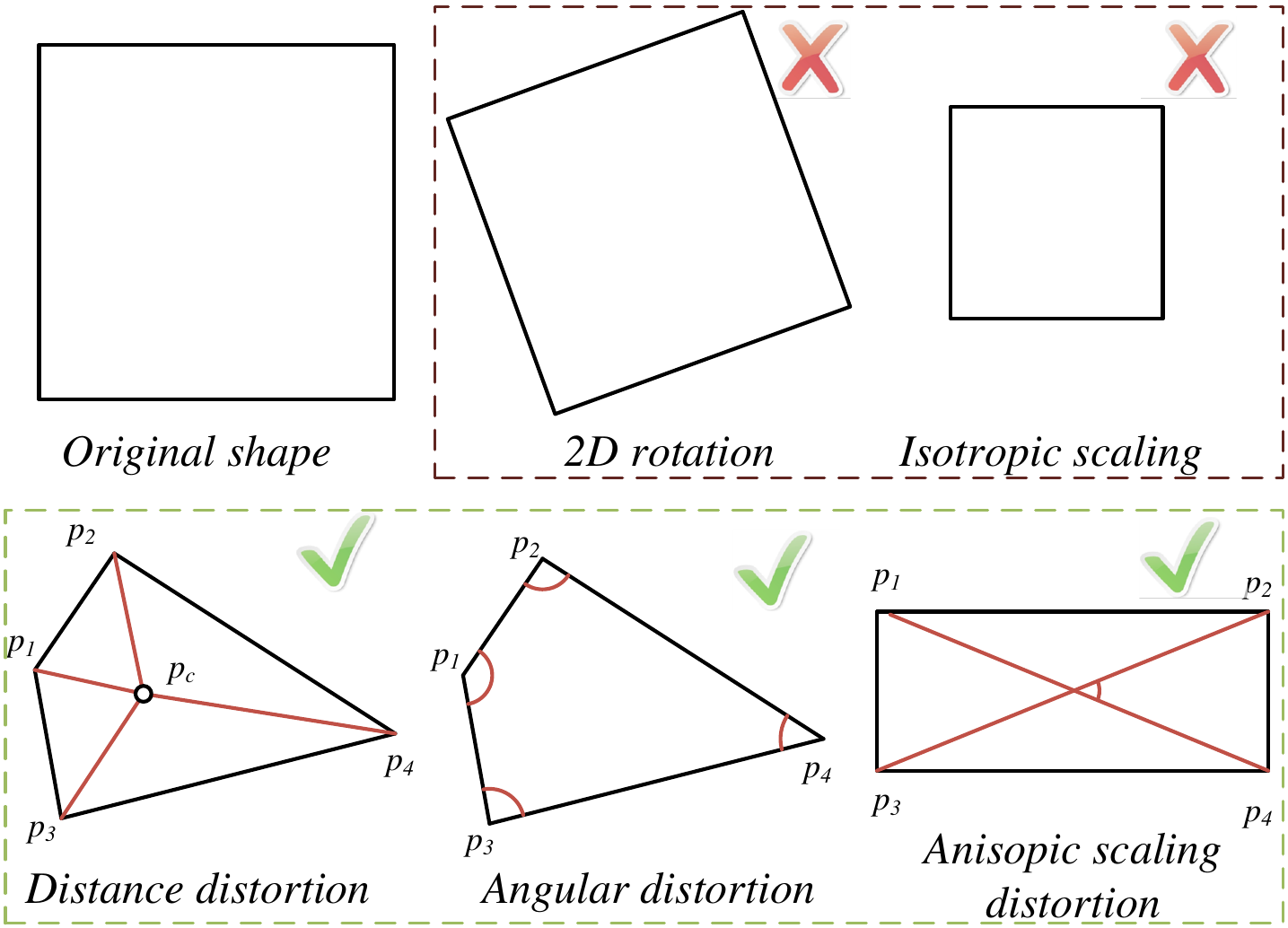} 
  \caption{Distortion categories. We measure the distortion degrees beyond the similarity transformation.} 
  \label{fig:distortion}
\end{figure}

 \subsubsection{Minimal Semantic Distortion}
To ensure the stitched image contains minimal semantic distortion, we define the distortion distribution map (DDM) and semantic distribution map (SDM). For the distortion distribution, we hold that a similarity transformation would not produce content distortion, including 2D rotation and isotropic scaling, as shown in Fig. \ref{fig:distortion}(top). 
To measure the deformation beyond the similarity transformation, we take a brief analysis of the difference between the homography and similarity transformations first. The similarity transformation has four DoF, including one for rotation, two for translations, and one for scale. The homography offers four additional degrees of freedom beyond those of a similarity transformation, including one for rotation, one for scale, and two for lines at infinity. Compared with the similarity transformation, the extra DoF of homography brings extra deformation regarding distance, angle, and overall scale (\textit{i.e}, extra rotation affects angles, anisotropic scale affects aspect ratio, and lines at infinity alter both angles and distances).

Hence, we measure the deformation distortion from the above three aspects. In particular, we define three types of distortion scores and assign them to each vertex, as shown in Fig. \ref{fig:distortion}(bottom). The first type is the \textbf{distance distortion score}, which can be calculated as follows:
\begin{equation}
    s_i^{distance} = \frac{\begin{vmatrix}\overrightarrow{p_ip_c}\end{vmatrix}}{min\{\begin{vmatrix}\overrightarrow{p_ip_c}\end{vmatrix}|i=1,2,3,4\}}-1,
\end{equation}
where $p_{i}$ represents the vertex and $p_{c}$ is the center of four vertece. The second type is the \textbf{angular distortion score}, as measured below by the cosine similarity:
\begin{equation}
    s_1^{angle} = \frac{\begin{vmatrix}\overrightarrow{p_1p_2}\cdot\overrightarrow{p_1p_3}\end{vmatrix}}{\begin{vmatrix}\overrightarrow{p_1p_2}\end{vmatrix}\cdot\begin{vmatrix}\overrightarrow{p_1p_3}\end{vmatrix}}.
\end{equation}
Here we calculate $p_1$'s score with its neighboring points as an example; the same regulation goes well for other vertices. The third type is the \textbf{global distortion score}, which is used to describe the magnitude of anisotropic scaling:
\begin{equation}
    s^{global} = \frac{\begin{vmatrix}\overrightarrow{p_1p_4}\cdot\overrightarrow{p_2p_3}\end{vmatrix}}{\begin{vmatrix}\overrightarrow{p_1p_4}\end{vmatrix}\cdot\begin{vmatrix}\overrightarrow{p_2p_3}\end{vmatrix}},
\end{equation}
where $\overrightarrow{p_1p_4}$ and $\overrightarrow{p_2p_3}$ are diagnals. The \textbf{final distortion score} is concluded by associating the three types of distortions as follows:
\begin{equation}
    s_i^{final} = s_i^{distance}+s_i^{angle} + s^{global}.
    \label{eq:dis_score}
\end{equation}
With each vertex being assigned a distortion score as Eq.~\eqref{eq:dis_score}, we can grid-sample pixel-wise distortion distribution maps ($D_{ref}$, $D_{tgt}$) for each view by bilinear interpolation.

For semantic distribution, the proposed constraint is compatible with any pretrained model capable of producing semantic feature maps.
After capturing the feature maps, we can compute the maximum and minimum values across two views, then apply min-max normalization to generate semantic distribution maps ($S_{ref}$,$S_{tgt}$).

By aligning DDM and SDM with the same resolution, we define the \textbf{semantic distortion loss} for our coefficient generator as:
\begin{equation}
    \mathcal{L}_{coef} = \Vert{\small D_{ref}\odot S_{ref}}\Vert_1 + \Vert{\small D_{tgt}\odot S_{tgt}}\Vert_1.
    \label{eq:sdl}
\end{equation}
For the iterative process, $D_{ref}$ and $D_{tgt}$ varies as $C_{dec}$ updates. So we encourage all iterations to carry out the constraint of minimal semantic distortion and set an exponentially decaying ratio ($0<\gamma<1$) to balance the significance of each iteration. Denoting the distortion distribution maps of each iteration as $D_{ref}^i$ and $D_{tgt}^i$, Eq. \eqref{eq:sdl} can be rewitten as:
\begin{equation}
    \mathcal{L}_{coef} =\sum_{i=1}^n \gamma^i(\Vert{\small D_{ref}^i\odot S_{ref}}\Vert_1 + \Vert{\small D_{tgt}^i\odot S_{tgt}}\Vert_1).
    \label{eq:sd2}
\end{equation}

Actually, the introduction of $\mathcal{L}_{coef}$ will inevitably degrade the warping performance. To handle this conflict, we propose a \textbf{two-stage training scheme}. For the first stage, we train the warping model with each decomposition coefficient being an independent random value between 0 and 1. After the warping model is enabled with the capability to align images with arbitrary coefficients, we freeze this part of parameters and only optimize the coefficient generator.

	\section{Experiment}
\label{section4}
\subsection{Dataset and Implementation Detail}
\subsubsection{Detail} 
We implemented our entire model using PyTorch and conducted all experiments on a single RTX 4090D GPU. The model training process is divided into two stages: the dual-branch registration network and the iterative coefficient generator. 
In the first stage, we only train the dual-branch alignment network with AdamW optimizer for 100 epochs. Both $\sigma$ and $C_{dec}$ parameters are randomly initialized to values between 0 and 1 for each batch. This strategy enhances the model's robustness against different cost volumes and planes. 
In the second stage, we freeze the parameters of the dual-branch alignment network and only train the iterative coefficient generation network for 50 epochs with $\sigma$ randomly sampled from $[0, 1]$. As for the minimal semantic distortion constraint, we use widely-adopted VGG19\cite{simonyan2014very} as the semantic feature extractor, demonstrating its versatility across common models. During inference, the average stitching time is measured to be 39ms per example on the UDIS-D dataset.

\subsubsection{Dataset} 
The datasets used in our experiments consist of two main parts: the UDIS-D dataset and a collection of classic image stitching datasets. The UDIS-D dataset contains 10,440 pairs of training samples and 1,106 pairs of testing samples. The classic image stitching dataset comprises 147 pairs of images across a variety of scenes, primarily sourced from \cite{lin2015adaptive,chang2014shape,chen2016natural,li2017parallax,herrmann2018object}. It covers a wide variety of stitching scenarios, including varying parallax, moving objects, significant illumination differences, and different aspect ratios. 

\subsection{Evaluation Metric} 
Prior works~\cite{nie2023parallax,nie2025stabstitch++,mei2024dunhuangstitch} typically adopt PSNR and SSIM as the evaluation metrics. However, these metrics are computed across the entire image, including invalid pixels after warping, which causes the objective measurements to be biased by the size of invalid regions.
Considering that, we adopt masked SSIM and masked PSNR (denoted as mSSIM and mPSNR) as quantitative metrics, which means we compute the average SSIM and PSNR values only for the valid overlapping pixels. 
Denoting $O_{ref}$ and $O_{tgt}$ as the warped images on the final canvas that can further be blended to get the final result $O$, we describe the calculation process as: 
\begin{equation}
  \begin{matrix}
    \begin{aligned}
     mSSIM = &\frac{\sum_{p \in O}M_{olp}(p) \cdot SSIM(p) }{ \sum_{p \in O}M_{olp}(p)},\\
     mPSNR = &20 \times log_{10}(\frac{1}{mRMSE}), \quad where \\
     mRMSE = &\frac{\sum_{p \in O}M_{olp}(p) \cdot (O_{ref}(p)- O_{tgt}(p))^2}{ \sum_{p \in O}M_{olp}(p)}.
    \end{aligned}
  \end{matrix}
\end{equation}
$M_{olp}$ is a 0-1 mask to identify the overlapping region, in which 1 indicates the pixel is overlapped. $p$ is the 2D coordinates in $O$, and $SSIM(\cdot)$ is computed using a window size of 7.





\subsection{Comparative Experiment}
To comprehensively evaluate the stitching performance of the proposed algorithm, we conduct a comparative analysis involving 7 stitching algorithms. These include 3 traditional stitching algorithms based on local mesh warping: APAP \cite{zaragoza2013projective}, SPW \cite{liao2019single}, and LPC \cite{jia2021leveraging}; 2 learning-based stitching algorithms leveraging global homography: UDIS \cite{nie2021unsupervised} and DunHuangStitch \cite{mei2024dunhuangstitch}; and 2 learning-based stitching algorithms employing TPS warping: UDIS++ \cite{nie2023parallax} and StabStitch++ \cite{nie2025stabstitch++}. 
StabStitch++ is a video stitching framework, in which the spatial warp is adopted to carry out comparisons with the same number of control points as ours. 
To visualize the alignment performance, all algorithms adopt average fusion for fairness.

\begin{table*}[h!]
	\centering
	\caption{Quantitative comparison on the UDIS-D dataset.}
	\label{tab1}
	\renewcommand{\arraystretch}{1.0}
	\setlength{\tabcolsep}{4pt}
    \scalebox{0.99}{
	\begin{tabular}{ccccccc|cccc}
		\hline
		 & \multirow{2}{*}{Method} & \multirow{2}{*}{Reference} & \multicolumn{4}{c}{mPSNR$\uparrow$} &  \multicolumn{4}{c}{mSSIM$\uparrow$}\\
		\cline{4-11} &  & & Easy & Moderate & Hard & Average & Easy & Moderate & Hard & Average \\
		\hline
		1 & APAP \cite{zaragoza2013projective} & CVPR2013 &26.77 & 22.88 & 18.75 & 22.39   &  0.868 & 0.770 & 0.587 & 0.726 \\
		2 &   SPW \cite{liao2019single} & TIP2019 & 25.82 & 21.49 & 15.85 & 20.52 &  0.844 & 0.693 & 0.434 & 0.634 \\
		3 & LPC \cite{jia2021leveraging} & CVPR2021 &  25.01 & 21.27 & 17.34 & 20.82 & 0.815 & 0.673  &0.485 & 0.640 \\
		\hline
		4 & UDIS \cite{nie2021unsupervised} & TIP2021&  23.53 & 19.73 & 17.42 & 19.94 & 0.761 & 0.545 & 0.376 & 0.542 \\
		5 & UDIS++ \cite{nie2023parallax}  & ICCV2023 &   27.58 & 23.75 & 20.04 & 23.41 &  0.880 & 0.792 & 0.632 & 0.755\\
		6 & DunHuangStitch \cite{mei2024dunhuangstitch} & TVCG2024 & 27.19 & 23.05 & 19.10 & 22.61 & 0.875 & 0.767 & 0.564 & 0.718 \\
		7 & StabStitch++ \cite{nie2025stabstitch++}  & TPAMI2025 &  29.92 & 24.93  & 20.46 & 24.63 &  0.927 & 0.845  & 0.664 & 0.797 \\
        8 & Ours & -  & 29.93 & 24.96 & 20.60 & 24.70 & 0.926 & 0.845 & 0.672 & 0.800\\
		\hline
		
	\end{tabular}
    }
    \vspace{0.1cm}
    \begin{tablenotes}
     \item {\scriptsize ‘Easy’, ‘Moderate’, and ‘Hard’  refer to the top 30\%, middle 30\%, and bottom 40\% of samples, according to performance rankings on the UDIS-D dataset.}
   \end{tablenotes}
\end{table*}

\begin{table*}[h!]
	\centering
	\caption{Quantitative comparison on classical datasets. $*$ denotes post processing with the iterative adaptation~\cite{nie2023parallax}.}
	\label{tab2}
	\renewcommand{\arraystretch}{1.0}
	\setlength{\tabcolsep}{4pt}
     \scalebox{0.99}{
	\begin{tabular}{ccccccc|cccc}
		\hline
		 & \multirow{2}{*}{Method} & \multirow{2}{*}{Reference}  & \multicolumn{4}{c}{mPSNR$\uparrow$} &  \multicolumn{4}{c}{mSSIM$\uparrow$}\\
		\cline{4-11} &  & &  Easy & Moderate & Hard & Average & Easy & Moderate & Hard & Average \\
		\hline
        1 & APAP \cite{zaragoza2013projective}& CVPR2013 & 24.33 & 19.48 & 14.47 & 18.92 & 0.818 & 0.687 & 0.442 & 0.628 \\
        2 & SPW\cite{liao2019single}& TIP2019 & 23.83 & 19.16 & 14.66 & 18.75 & 0.793 &  0.647 & 0.445 &  0.610 \\   
        3 & LPC \cite{jia2021leveraging} & CVPR2021& 23.19 & 18.67 & 13.90 & 18.11 & 0.761 & 0.600 & 0.412 & 0.573 \\
        \hline
        4 & UDIS \cite{nie2021unsupervised} & TIP2021& 20.58 & 16.27 & 13.37 & 16.40 & 0.660 & 0.471 & 0.271& 0.454 \\
	5 & UDIS++ \cite{nie2023parallax}  & ICCV2023 &  21.93 &  17.53 & 13.83 & 17.36 & 0.709  &  0.526 & 0.325 &  0.500 \\
        6 & DunHuangStitch \cite{mei2024dunhuangstitch}& TVCG2024 & 22.02 & 17.44 & 13.65 & 17.29 & 0.712 & 0.534 & 0.319 & 0.501 \\
        7 & StabStitch++  \cite{nie2025stabstitch++} & TPAMI2025 & 23.01  & 18.08  & 13.98 & 17.91 &  0.751 & 0.577  & 0.339 &  0.534 \\
        8 & Ours  & - & 23.40 & 18.54 & 14.67 & 18.44 & 0.772  & 0.607 & 0.387 & 0.568  \\
         9 & Ours$^{*}$ & - &25.40 & 19.79 & 15.48 & 19.74 & 0.839 & 0.691 & 0.465 & 0.645 \\
		\hline
	\end{tabular}
    }
    \vspace{0.1cm}
        \begin{tablenotes}
     \item {\scriptsize ‘Easy’, ‘Moderate’, and ‘Hard’  refer to the top 30\%, middle 30\%, and bottom 40\% of samples, according to performance rankings on classic datasets.}
   \end{tablenotes}
\end{table*}

\subsubsection{Quantitative Comparison}

Tables \ref{tab1} and \ref{tab2} respectively present the alignment performance of various stitching algorithms on the UDIS-D dataset and classical stitching datasets. As shown in Table \ref{tab1}, among traditional stitching algorithms, APAP demonstrates better alignment performance. This is primarily because SPW and LPC impose more constraints to enhance content fidelity, which in turn weakens their alignment capabilities. UDIS and DunHuangStitch, both trained on 128-resolution images and employing a three-stage cascaded global alignment strategy, perform significantly worse than local mesh warping-based algorithms like UDIS++ and StabStitch++, especially on the scenes of parallax.

From Table \ref{tab2}, it can be observed that, due to the robust prior knowledge of feature point detection, traditional stitching algorithms exhibit clear advantages in cross-scene stitching compared to learning-based algorithms. This highlights a critical limitation of learning-based stitching algorithms: their significant cross-scene restrictions and weak generalization capabilities. However, as shown in row 8, the proposed algorithm demonstrates remarkable cross-scene generalization performance, with its zero-shot capability clearly surpassing other learning-based stitching algorithms.

Although the zero-shot performance of the proposed algorithm still lags behind traditional stitching algorithms, we can further enhance its alignment performance using the iterative adaptation strategy proposed in UDIS++. This result indicates that the proposed algorithm not only possesses robust zero-shot cross-scene generalization capabilities but also exhibits stable data-fitting potential.

\subsubsection{Qualitative Comparison}
\begin{figure*}[h!]
	\centering
	\includegraphics[width=\linewidth]{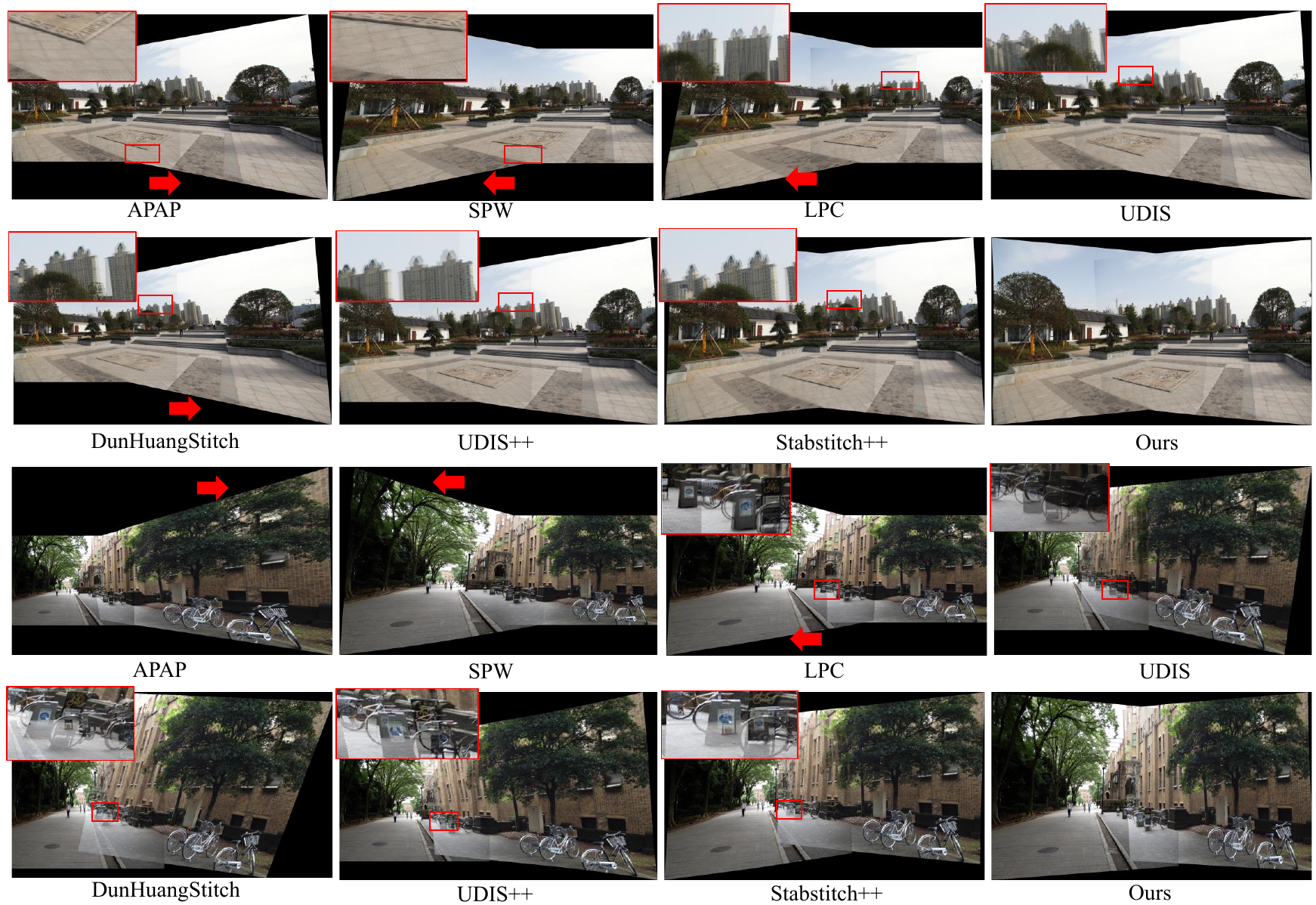}
	\caption{Performance comparison on classical datasets. Arrows indicate regions with noticeable stretching, while rectangular boxes highlight areas with significant content misalignment.} 
	\label{fig:exp_fig3}
\end{figure*}

Here, we only demonstrate the visualization results on classic datasets, with a special highlight on the robustness and naturalness.
As illustrated in Fig. \ref{fig:exp_fig3}, LPC algorithm exhibits significant content stretching issues, while UDIS++ tends to produce more background gaps and partial content misalignment. Both of them get abundant invalid pixels because both of them adopt the single-view warp to align images. 
StabStitch++ employs an intermediate plane and achieves relatively natural stitching results. But its performance on cross-dataset scenes, especially the alignment capability, is not totally satisfactory. In contrast, the proposed algorithm strikes a balance between content alignment and the naturalness of the stitched results, exhibiting neither obvious content stretching nor excessive background gaps. Our superior visual quality is attributed not only to the use of a dual-branch backbone, which introduces universal prior knowledge and enhances the network’s cross-domain generalization ability, but also to the incorporation of the optimal plane, which ensures the stitching naturalness.

\subsubsection{Zero-shot Image Stitching}
\begin{figure*}[htbp!] 
	\centering
	\includegraphics[width=\linewidth]{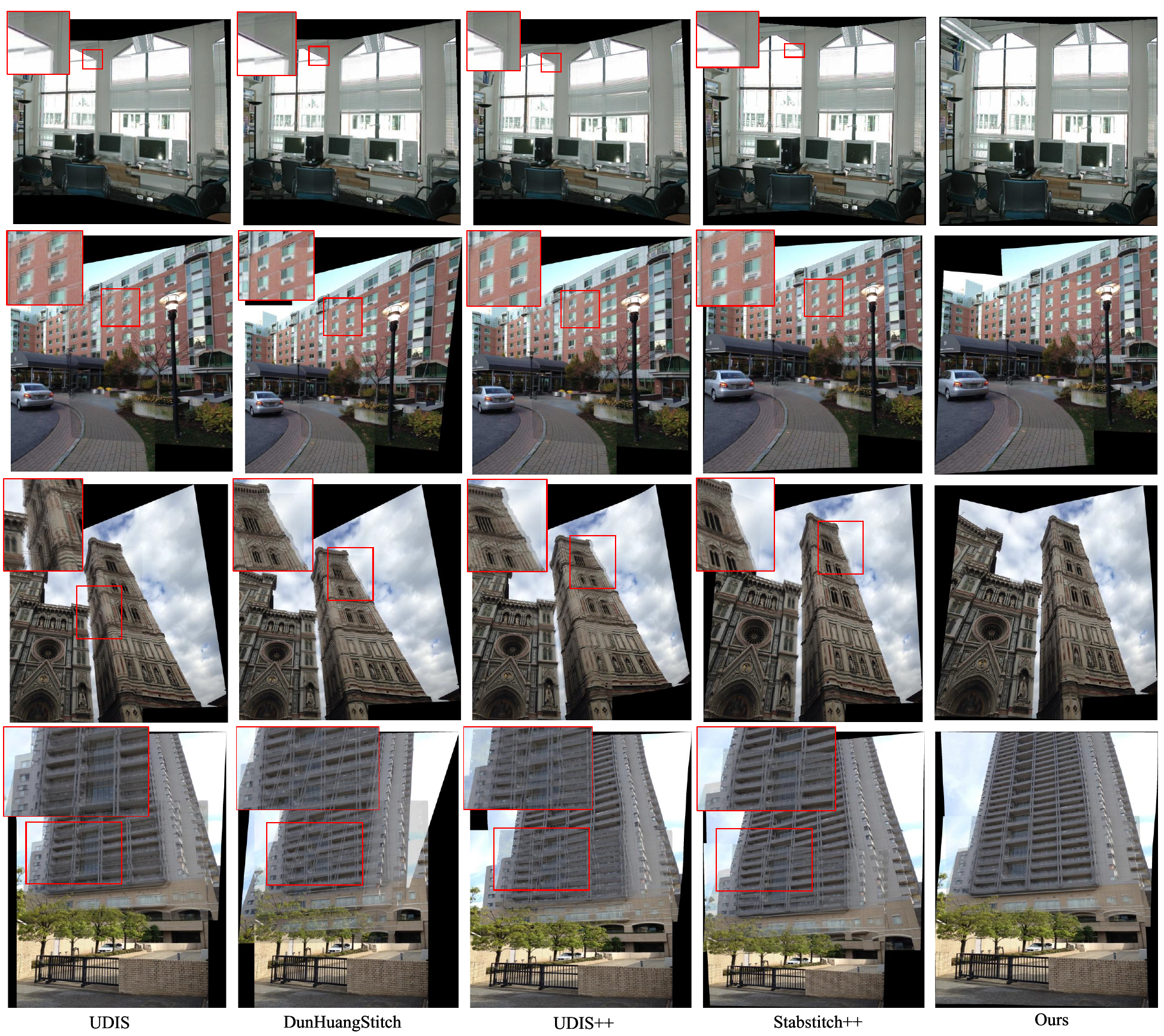}
	\caption{Zero-shot comparative results of learning-based stitching algorithms.}  
	\label{fig:zero-shot}
\end{figure*}

Unlike previous qualitative comparisons, this study highlights the zero-shot image stitching performance of state-of-the-art learning-based stitching solutions. In Fig. \ref{fig:zero-shot}, all methods are trained on the UDIS-D dataset, but tested on unseen scenes.
The proposed algorithm demonstrates superior zero-shot generalization performance across multiple scenes compared to existing learning-based stitching algorithms, yielding better content alignment results and more natural structures simultaneously.

\begin{figure*}[htbp!] 
	\centering
	\includegraphics[width=\linewidth]{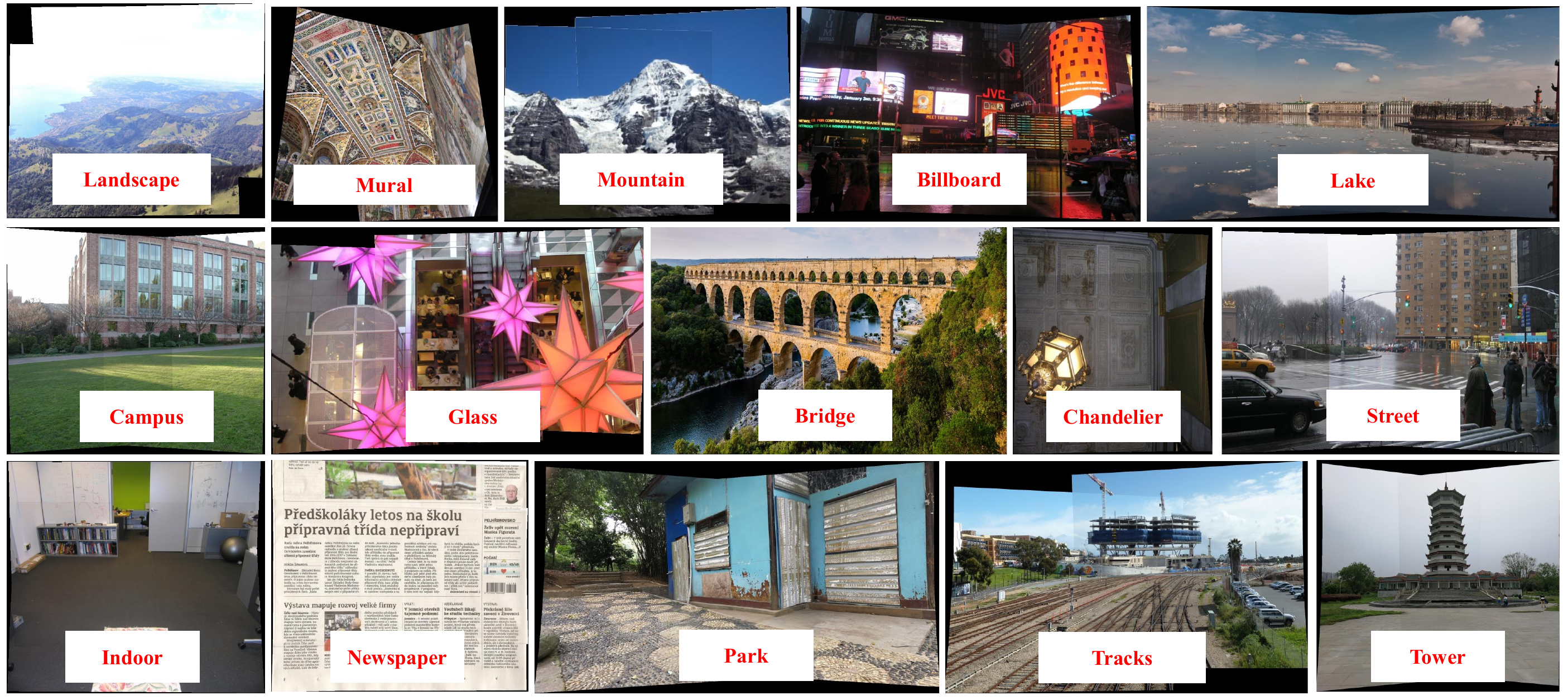}
	\caption{More cross-domain results of the proposed solution in a wide diversity of scenes. } 
	\label{fig:multi_scence}
\end{figure*}

In addition to these comparisons, we demonstrate more stitching results across a wide diversity of scenes in Fig. \ref{fig:multi_scence}. All instances are sourced from the universally acknowledged and widely used classic datasets. Our algorithm achieves robust and appealing stitching results across complex scenarios, including indoor and outdoor environments (\textit{e.g.}, Chandelier \cite{li2017parallax} and Park \cite{jia2021leveraging} ), dense and sparse objects (\textit{e.g.}, Newspaper \cite{brown2007automatic} and Lake \cite{chang2014shape}), and high-exposure and low-light conditions (\textit{e.g.}, Landscape \cite{nomura2007scene} and Billboard \cite{du2022geometric}), demonstrating strong cross-scenario generalization performance.


\subsection{ Ablation Study}

\subsubsection{Impact of Backbone}

Table \ref{tab3} presents the content alignment results obtained using different backbone architectures and whether the proposed dual-branch backbone is employed. The alignment performance on the UDIS-D dataset reflects the algorithm's generalization ability in similar scenarios, while the performance on classic datasets demonstrates its generalization across different scenarios. 
Comparing row 1 with row 2, we can observe that a single trainable branch demonstrates superior performance in scenarios with small domain gaps (\textit{i.e.}, UDIS-D), while a single frozen branch excels in large domain gap situations (\textit{i.e.}, classic datasets).
We can further notice that the proposed dual-branch architecture integrates both of their advantages, when comparing row 1 and 2 with row 3.
Besides, the performance can be further improved using our ternary search strategy, by analysing row 3 and row 4.
In addition to ResNet18, we also tried other backbones (\textit{e.g.}, ResNet50). Rows 5 to 8 reveal the same pattern, which exhibits the universality of our architecture.
Finally, we conduct more experiments with each branch adopting different backbones. 
As shown in rows 9 and 10,  we replace the frozen branch in row 3 with other backbones, such as ResNet50, and DINOv2-s~\cite{oquab2023dinov2}. The results demonstrate that leveraging the same network architectures for the two branches is conducive to constructing unified representations between the activated and frozen branches, yielding better stitching quality. 


\begin{table}[htbp!]
	\centering
	\caption{Ablation studies on different backbones.}
	\label{tab3}
	\renewcommand{\arraystretch}{1.0}
	\setlength{\tabcolsep}{2pt}
	\begin{tabular}{cccc|cc}
		\hline
		 &  \multirow{2}{*}{Search Strategy} & \multicolumn{2}{c}{Backbone}  & \multicolumn{2}{c}{mSSIM$\uparrow$}\\
        \cline{3-6} & &$f_{frozen}$ & $f_{train}$ & UDIS-D & Classic \\
		\hline
        1 & $\sigma=0$&- & ResNet18 & 0.800 & 0.525\\
        2 & $\sigma=1$&ResNet18 &- & 0.781 & 0.530\\
        3 & $\sigma=0.5$&ResNet18 & ResNet18   & 0.793 & 0.543\\
        4 & Ternary Search &ResNet18  & ResNet18 & 0.800 & 0.568\\
        \hline
        5 & $\sigma=0$&- & ResNet50  & 0.804 & 0.519\\
        6 & $\sigma=1$&ResNet50 & -  & 0.780 & 0.529 \\
        7 & $\sigma=0.5$&ResNet50 & ResNet50 & 0.799 & 0.542\\
        8 & Ternary Search &ResNet50 & ResNet50 & 0.804 & 0.565 \\
        \hline
        9 & $\sigma=0.5$&ResNet50 & ResNet18 & 0.795 & 0.535 \\
        10 & $\sigma=0.5$& DINOv2-s & ResNet18 & 0.786 & 0.523\\
        \hline

	\end{tabular}
\end{table}

\subsubsection{Impact of Aggregation Strategy}

Table \ref{tab4} evaluates the impact of different aggregation strategies (\textit{i.e.}, aggregation position and mode) on the model's generalization ability. We carry out the evaluation on the classic datasets without ternary search. From that, we can observe two insights. (1) Comparing row 1 and 2 with row 5 shows that integrating information at the feature dimension is less effective for cross-scenario generalization than integrating information on the correlation level. (2) Comparing row 3 and 4 with row 5 demonstrates that employing a random $\sigma$ strategy during training can significantly enhance the model’s cross-scenario generalization performance, effectively preventing the frozen and activated backbones from learning homogeneous feature representations.

\begin{table}[htbp!]
	\centering
	\caption{Ablation studies using different aggregation strategies during the training stage. }
	\label{tab4}
	\renewcommand{\arraystretch}{1.0}
	\setlength{\tabcolsep}{1pt}
    \scalebox{0.99}{
	\begin{tabular}{ccc|cc}
		\hline
        & \multicolumn{2}{c}{Aggregation Strategy} & \multicolumn{2}{c}{mSSIM$\uparrow$} \\
        \cline{2-5}
		  & Position & Mode  &UDIS-D &  Classic\\
		\hline
         1\quad\quad &  Feature Level ($F^{1/16}$)& Random & 0.797 & 0.537\\
       	2 \quad\quad& Feature Level($F^{1/16}$,$F^{1/8}$) & Random &  0.795 & 0.536 \\
        \hline
       3\quad\quad &  Correlation Level & Average & 0.793 & 0.525 \\
        4\quad\quad & Correlation Level& Learnable & 0.791 & 0.534\\
        \hline
        5\quad\quad &  Correlation Level& Random   & 0.793 & 0.543\\
		\hline
	\end{tabular}
    }
\end{table}

\subsubsection{Impact of Optimal Plane}

\begin{figure}[htbp!]
	\centering
	\includegraphics[width=\linewidth]{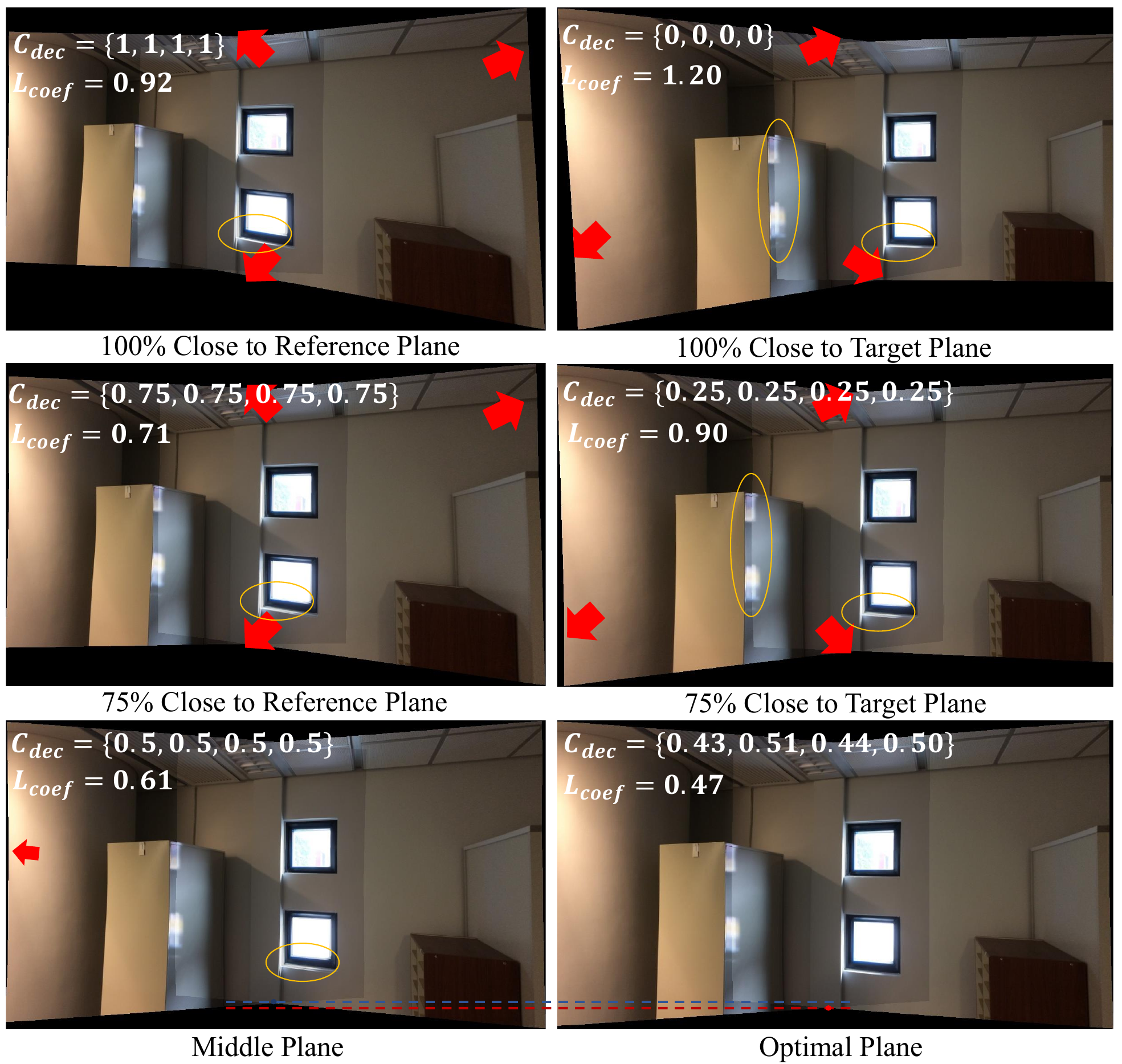}
	\caption{Comparison of stitching results based on different virtual planes. The red arrows highlight differences in the background areas, while the yellow circles indicate areas of content misalignment.} 
	\label{fig:exp_fig4}
\end{figure}

Fig. \ref{fig:exp_fig4} illustrates the differences between our optimal plane and other planes used for warping. It can be observed that aligning content using either the reference plane or the target plane results in obvious perspective distortions and substantial background invalid regions. As the stitching plane is gradually adjusted, the invalid area is gradually reduced, and distortion is accordingly relieved. 
Compared with these artificially controlled planes, the optimal plane is optimized with minimal semantic distortions, which preserve the semantically salient regions from unnatural deformation. 
Moreover, as shown in Table \ref{tab5}, transitioning the stitching plane from the reference plane to the intermediate plane and finally to our optimal plane improves naturalness (measured by $L_{coef}$) without compromising alignment performance (quantified by mSSIM). This mutually beneficial outcome is achieved through our carefully designed two-stage training scheme.


\begin{table}[htbp!]
	\centering
	\caption{The quantitative impact of optimal plane on the alignment and naturalness.}
	\label{tab5}
	\renewcommand{\arraystretch}{1.0}
	\setlength{\tabcolsep}{2pt}
    \scalebox{0.99}{
	\begin{tabular}{cccc|cc}
		\hline

          & \multirow{2}{*}{Plane}  & \multicolumn{2}{c}{mSSIM$\uparrow$} & \multicolumn{2}{c}{$L_{coef}$$\downarrow$} \\
          \cline{3-6}
          & & UDIS-D & Classic &  UDIS-D & Classic \\
		\hline
        1 &  Refence Plane  & 0.786 & 0.537 &  0.48 & 0.79\\
         2 &  Middle Plane & 0.800  & 0.569 & 0.26 & 0.49\\
        3  & Optimal Plane (ours) & 0.800  & 0.568 & 0.23 & 0.42\\
		\hline
	\end{tabular}
    }
\end{table}

	\section{Conclusion}

In this paper, we propose \textit{RopStitch}, a robust and natural image stitching framework. It distinguishes itself from existing solutions through two major contributions.
First, it introduces the universal prior of content perception through a dual-branch architecture. This architecture simultaneously aggregates semantically invariant features and fine-grained discriminative features, facilitating cross-domain generalization performance. Then, we present to stitch images on a virtual optimal plane to reduce structure distortions. It redistributes the warping burdens of respective views at the principle of minimal semantic distortion without sacrificing alignment performance.
Experiments across diverse datasets exhibit our superiority over existing methods and validate the universality of its components.

    \bibliographystyle{IEEEtran}
	\bibliography{bib}


\begin{IEEEbiography}[{\includegraphics[width=1in,height=1.25in,clip,keepaspectratio]{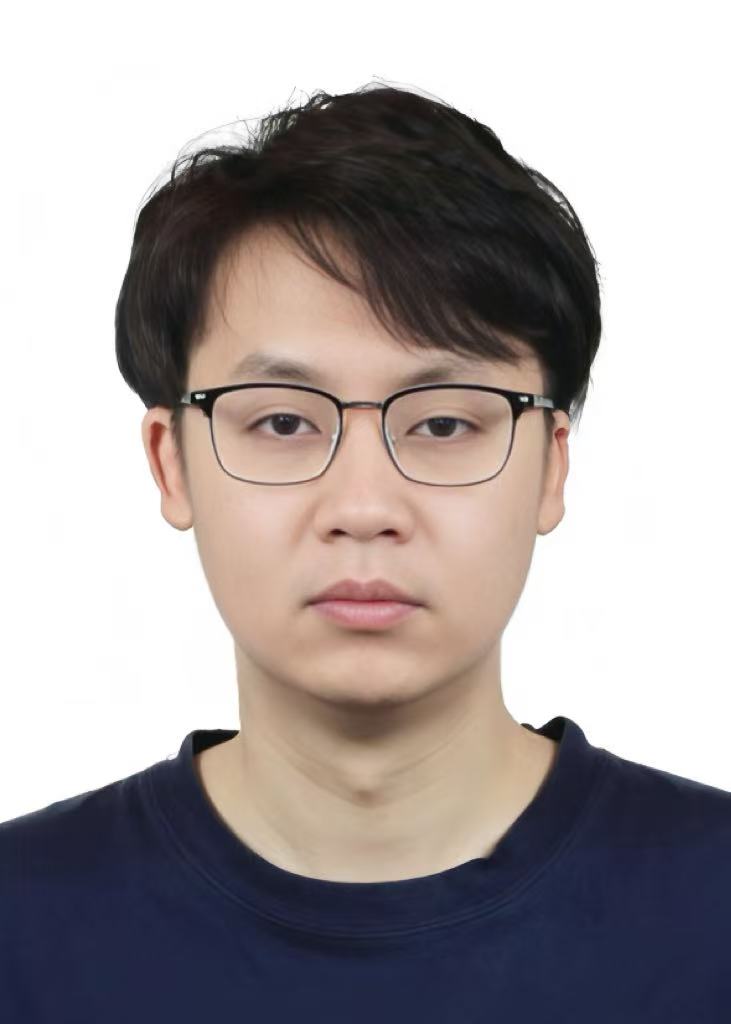}}]{Lang Nie}(Member, IEEE) is currently an Associate Professor at Chongqing University of Posts and Telecommunications, China. Prior to that, he received his B.S. and Ph.D. degrees from the School of Computer Science and Technology, Beijing Jiaotong University (BJTU) in 2019 and 2025. His current research interests include image and video processing, 3-D vision, and multi-view geometry.
\end{IEEEbiography}

\begin{IEEEbiography}[{\includegraphics[width=1in,height=1.25in,clip,keepaspectratio]{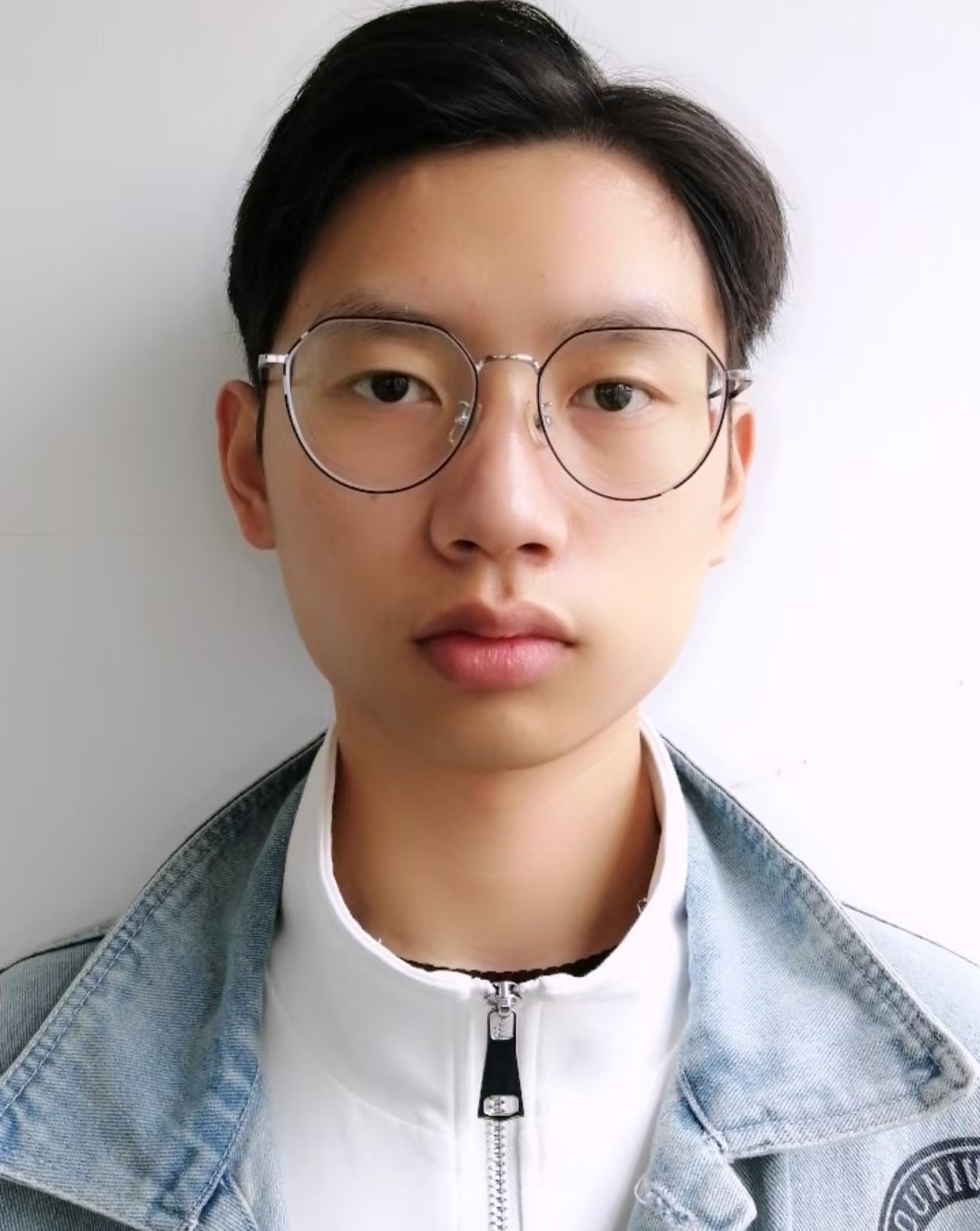}}]{Yuan Mei} is currently a PH.D. student at The Hong Kong Polytechnic University. His research interests include computer vision, multimodal information processing, and vision and language navigation.
\end{IEEEbiography}

\begin{IEEEbiography}[{\includegraphics[width=1in,height=1.25in,clip,keepaspectratio]{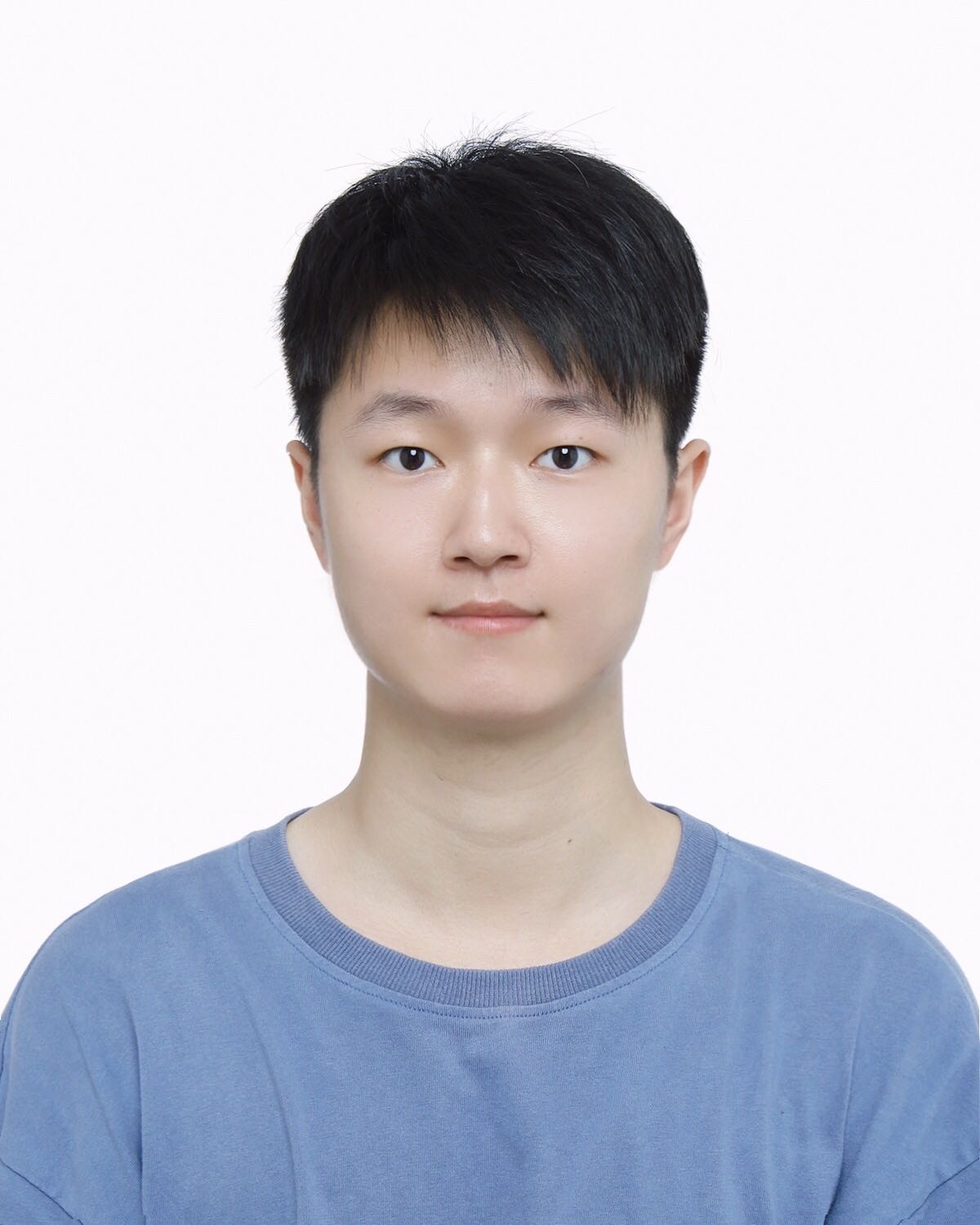}}]{Kang Liao} (Member, IEEE) is currently a Research Fellow at Nanyang Technological University (NTU), Singapore. Before joining NTU, he received his Ph.D. degree from Beijing Jiaotong University in 2023. From 2021 to 2022, he was a Visiting Researcher at Max Planck Institute for Informatics in Germany. His current research interests include camera calibration, 3D vision, and panoramic vision.
\end{IEEEbiography}

\begin{IEEEbiography}[{\includegraphics[width=1in,height=1.25in,clip,keepaspectratio]{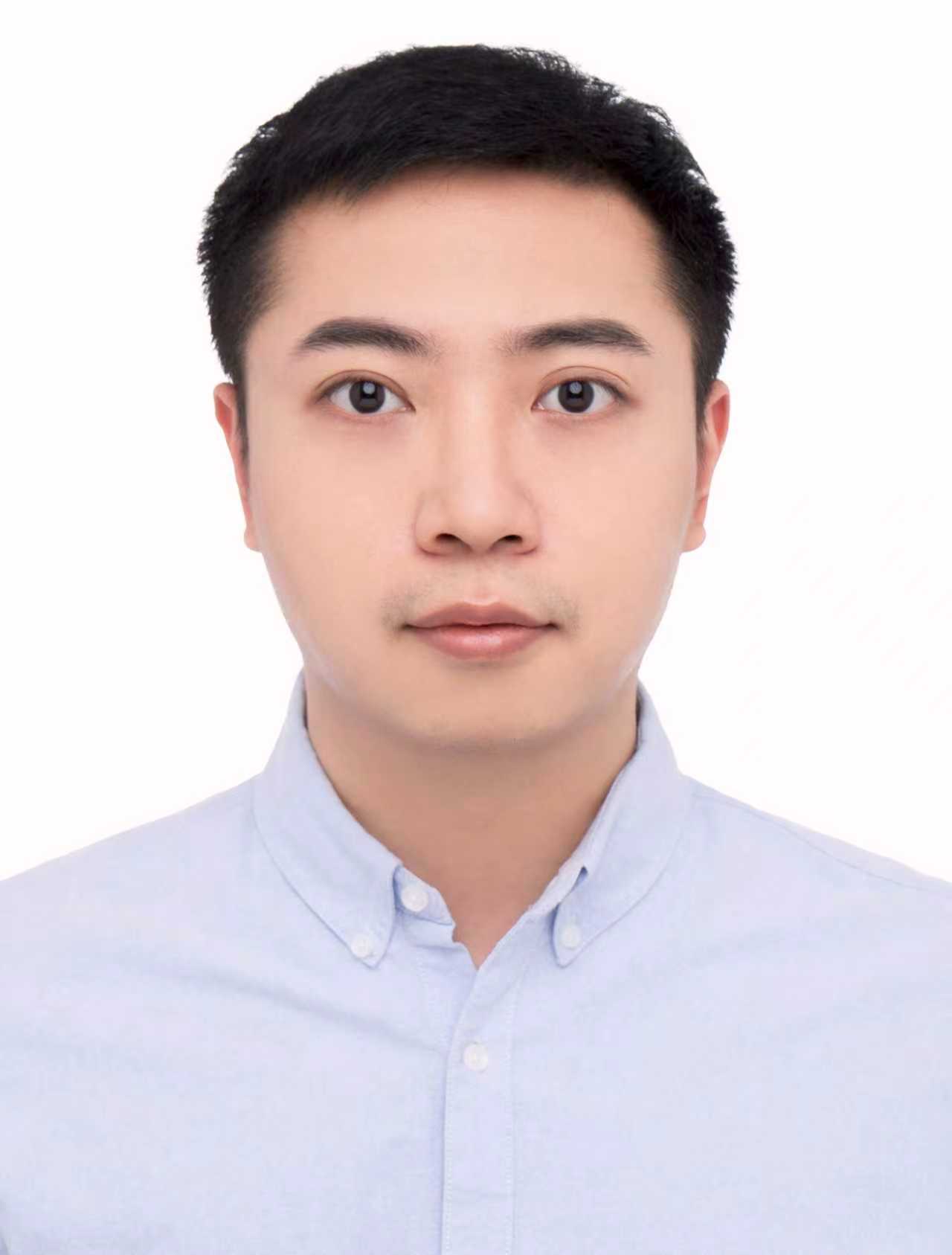}}]{Yunqiu Xu} (Member, IEEE) is currently a Postdoctoral Researcher at Zhejiang University. He received the Ph.D. degree from the University of Technology Sydney in 2023. Prior to that, he received the M.E. and B.E. degrees from Chongqing University of Posts and Telecommunications. His research interests include computer vision, multimodal learning, and weakly supervised learning.
\end{IEEEbiography}

\begin{IEEEbiography}[{\includegraphics[width=1in,height=1.25in,clip,keepaspectratio]{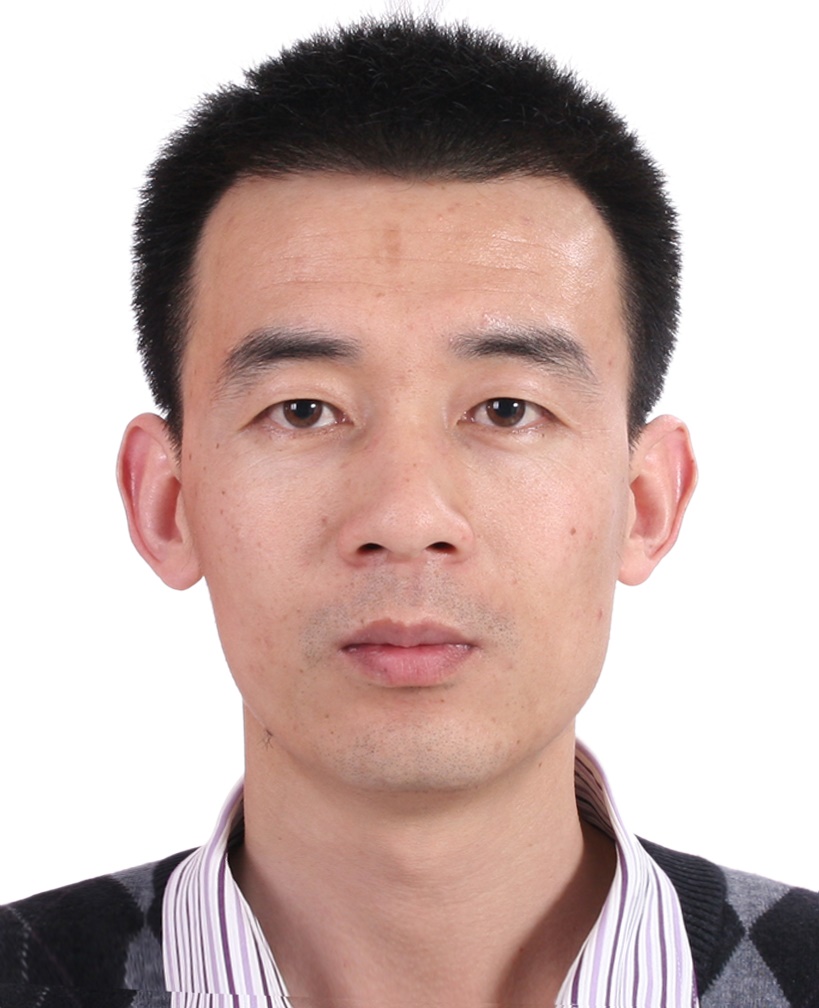}}]{Chunyu Lin}(Member, IEEE) is a Professor at Beijing Jiaotong University. He received the Ph.D. degree from Beijing Jiaotong University (BJTU), Beijing, China, in 2011. From 2009 to 2010, he was a Visiting Researcher at the ICT Group, Delft University of Technology, Netherlands. From 2011 to 2012, he was a Post-Doctoral Researcher with the Multimedia Laboratory, Gent University, Belgium. His research interests include multi-view geometry, camera calibration, and virtual reality video processing.
\end{IEEEbiography}

\begin{IEEEbiography}[{\includegraphics[width=1in,height=1.25in,clip,keepaspectratio]{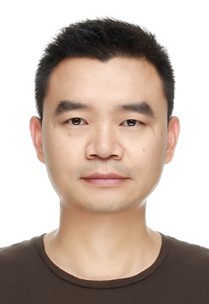}}]{Bin Xiao} (Senior Member, IEEE) is a Professor at Chongqing University of Posts and Telecommunications, Chongqing, China. He received the Ph.D. degree in computer science from Xidian University, Xi’an, China, in 2012. His research interests include image processing, pattern recognition, and digital watermarking.
\end{IEEEbiography}


\vfill

\end{document}